\documentclass[conference]{IEEEtran}
\IEEEoverridecommandlockouts
\usepackage{cite}
\usepackage{amsmath,amssymb,amsfonts}
\usepackage{graphicx}
\usepackage{textcomp}
\usepackage{xcolor}
\usepackage{booktabs}
\usepackage{multirow}
\usepackage{url}

\begin{document}

\title{Geometric Filtering of LLM-Generated Samples for Few-Shot Text Classification


}

\author{\IEEEauthorblockN{Benjam\'{i}n Schindler}
\IEEEauthorblockA{\textit{Faculty of Engineering and Sciences} \\
\textit{Universidad Adolfo Ib\'{a}\~{n}ez}\\
\textit{Millennium Nucleus for Social Data Science (SODAS)}\\
Santiago, Chile \\
bschindler@alumnos.uai.cl}
\and
\IEEEauthorblockN{Gonzalo A. Ruz}
\IEEEauthorblockA{\textit{Faculty of Engineering and Sciences} \\
\textit{Universidad Adolfo Ib\'{a}\~{n}ez}\\
\textit{Millennium Nucleus for Social Data Science (SODAS)}\\
\textit{Millennium Nucleus in Data Science for}\\
\textit{Plant Resilience (PhytoLearning)}\\
Santiago, Chile \\
gonzalo.ruz@uai.cl}
}

\maketitle


\begin{abstract}
Large language models (LLMs) can generate synthetic training data for text classification, but the quality of generated samples is heterogeneous: some fall in correct class regions of the embedding space while others land in peripheral or cross-class zones. We propose a geometric filtering framework that evaluates each LLM-generated sample by its Euclidean distance to real class examples in a sentence embedding space, selecting only geometrically consistent candidates. A soft weighting mechanism transforms filter scores into sample weights for classifier training. Evaluated across 13 datasets, 5 classifiers, 10 augmentation methods, and over 6,700 configurations, our method achieves +2.61 percentage points (pp) over SMOTE ($p<0.0001$, Cohen's $d=0.95$, 88.9\% win rate). The approach generalizes to named entity recognition (+9.26pp, 100\% win rate) without filter modification, and is robust across 5 LLMs from 4 providers. A key finding is that the simplest distance-based filter consistently outperforms complex multi-criteria alternatives.
\end{abstract}

\begin{IEEEkeywords}
data augmentation, large language models, few-shot learning, text classification, geometric filtering, embedding space
\end{IEEEkeywords}

\section{Introduction}

Text classification in low-resource scenarios remains a central challenge in natural language processing. Many practical domains, such as hate speech detection, news classification, and spam filtering, require supervised models trained on labeled data, yet practitioners often have only 10 to 50 labeled examples per class. This scarcity is compounded in multi-class settings, where each category needs a critical mass of examples for the classifier to learn robust decision boundaries. The consequence is degraded macro F1 performance, particularly for underrepresented classes~\cite{gopali2024applicability}.

Two main paradigms address this data scarcity through augmentation. \textit{Classical oversampling} methods such as SMOTE~\cite{chawla2002smote} and ADASYN~\cite{he2008adasyn} generate synthetic samples via interpolation in the embedding space. These methods provide explicit geometric guidance on sample placement but produce interpolated vectors that do not correspond to linguistically coherent text. \textit{LLM-based generation} produces grammatically correct and semantically plausible text~\cite{whitehouse2023llm,brown2020fewshot}, but lacks information about the geometric structure of the representation space where classifiers operate.

A fundamental gap separates these paradigms: classical methods offer geometric control without linguistic validity, while LLMs provide linguistic quality without geometric awareness. This creates a quality heterogeneity problem: LLM-generated samples vary widely in their embedding space placement. Some land in regions faithful to the target class, while others fall in peripheral zones or even in regions associated with other classes. Incorporating all generated samples indiscriminately can degrade classifier performance~\cite{li2023synthetic,cegin2025llms}. Springer et al.~\cite{springer2025influence} further demonstrate that the influence of synthetic data on few-shot classifiers depends critically on sample quality, not just quantity. This observation motivates our central question: \textit{can LLM-based augmentation be improved through geometric filtering that selects only high-quality synthetic samples?}

This paper proposes a post-generation geometric filtering system that evaluates each synthetic sample by its position in the embedding space. The framework bridges the two paradigms: LLMs generate linguistically valid text, and geometric filters ensure that only samples consistent with the real data distribution are retained. Our contributions are:

\begin{enumerate}
    \item A geometric filtering framework based on Euclidean distance in the embedding space that selects high-quality LLM-generated samples. The system is decoupled from the generative model and validated with 5 LLMs from 4 providers.
    \item A soft weighting mechanism that transforms geometric filter scores into continuous sample weights for classifier training, going beyond binary accept/reject decisions.
    \item Demonstration that these filters generalize across tasks, from text classification to named entity recognition (NER), without modification, achieving +9.26pp improvement.
    \item Comprehensive empirical evidence across 6,700+ configurations that the simplest distance-based filter consistently outperforms complex multi-criteria alternatives, with theoretical analysis explaining why.
\end{enumerate}

\section{Related Work}

\subsection{Classical Oversampling}

SMOTE~\cite{chawla2002smote} generates synthetic samples by interpolating between nearest neighbors in feature space. ADASYN~\cite{he2008adasyn} extends this with adaptive sampling density. EDA~\cite{wei2019eda} applies surface-level text perturbations (synonym substitution, random insertion/deletion). While effective in certain settings, these methods either lack linguistic coherence (SMOTE in embedding space) or produce only superficial variations (EDA)~\cite{feng2021survey,taskiran2025comprehensive}.

\subsection{LLM-Based Augmentation}

GPT-3-generated reviews improve accuracy by +12.7\% and macro F1 by +13pp in sentiment analysis~\cite{suhaeni2023mitigating}. Fine-tuned LLMs outperform SMOTE for multiclass imbalanced text, achieving F1 of 0.76 vs 0.60~\cite{cloutier2023fine}. AugGPT leverages ChatGPT for augmentation across multiple tasks~\cite{dai2023auggpt}. However, recent work shows that LLM benefits diminish when more than 20 examples per class are available~\cite{cegin2025llms}, and that evaluation protocols significantly impact conclusions~\cite{piedboeuf2024evaluation}.

\subsection{Hybrid and Quality-Aware Approaches}

Emerging methods combine geometric and generative approaches. LLMOverTab uses SMOTE to identify underrepresented regions and LLMs to fill them~\cite{isomura2025llmovertab}. SMOTExT applies SMOTE interpolation followed by LLM decoding~\cite{smotext2025arxiv}. SentiGEN combines T5 with genetic algorithms and an XLNet validator~\cite{sundararajan2023sentigen}. AugmenToxic applies LLM augmentation for toxicity detection~\cite{bodaghi2024augmentoxic}. However, none of these methods implement quantitative post-generation validation in the embedding space. Our work addresses this gap by introducing geometric filters that evaluate sample quality based on embedding space properties.

\section{Methodology}

\subsection{Problem Setup}

We evaluate on 13 datasets spanning 9 textual domains: 7 text classification datasets (2--20 classes) including 20newsgroups, sms\_spam, hate\_speech, ag\_news, emotion, dbpedia14, and 20newsgroups\_20class; 3 scalability datasets (trec6, banking77, clinc150 with 6--150 classes); and 3 NER corpora (MultiNERD, WikiANN, Few-NERD). Few-shot scenarios are simulated by sampling 10, 25, or 50 examples per class from training sets, keeping full test sets intact.

All texts are encoded using \texttt{all-mpnet-base-v2}~\cite{reimers-gurevych-2019-sentence,muennighoff2023mteb}, producing L2-normalized 768-dimensional embeddings. An ablation with 3 additional embedding models (bge-large, e5-large, bge-small) confirms robustness (90.5\% cross-model agreement).

\subsection{LLM Generation}

We use Gemini 3 Flash as the primary generative model, with robustness validated across GPT-5-mini, Claude 4.5 Haiku, Kimi K2.5, and GLM-5 (5 models, 4 providers). For each class, real examples serve as in-context demonstrations and 3$\times$ surplus candidates are generated (3 candidates per desired sample). No fine-tuning is applied; all specificity comes from in-context learning. Responses are cached via MD5 hashing for reproducibility.

\subsection{Geometric Filters}

We implement five filtering strategies with increasing complexity:

\textbf{No filter (control):} Random selection from candidates, isolating the contribution of geometric filtering.

\textbf{LOF filter}~\cite{breunig2000lof}: Evaluates local density of each synthetic sample relative to real class examples. Two variants: relaxed (permissive threshold) and strict.

\textbf{Cascade filter:} A hierarchical scoring system with up to 4 levels: (1) Euclidean distance to nearest real anchor, (2) cosine similarity, (3) KNN purity, (4) centroid distance. Scores are combined via geometric mean and top-$N$ candidates are selected by ranking. \textbf{Level 1 (distance only) produces the best results.}

\textbf{Combined filter:} Requires passing both LOF and cosine similarity thresholds simultaneously. This is the most restrictive strategy, included as a case study of over-filtering.

The key finding is that level-1 cascade (Euclidean distance only) outperforms all multi-criteria alternatives. For L2-normalized vectors, $\|u-v\|_2^2 = 2\,(1-\cos(u,v))$, so Euclidean distance is a strictly monotonic function of cosine similarity and the latter adds no discriminative information~\cite{tessari2024diem}. Additionally, LOF requires $k=20$ neighbors but only 9 are available at 10-shot, producing degenerate density estimates~\cite{robinson2025manifold}.

\subsection{Soft Weighting}

Instead of binary accept/reject, geometric filter scores are transformed into continuous sample weights. The procedure has four steps: (1) compute cascade level-1 scores for all candidates; (2) select top-$N$ candidates; (3) normalize scores to $[0,1]$ via min-max; (4) apply temperature scaling:
\begin{equation}
w(x) = \max\left(w_{\min},\ s_{\text{norm}}(x)^{1/T}\right)
\label{eq:weight}
\end{equation}
where $T=0.5$ (i.e., an exponent of $1/T=2$, which sharpens the weight
distribution toward the highest-scoring candidates) and $w_{\min}=0$, so that the $\max$ operator in~(\ref{eq:weight}) is inactive here and is
retained only as a general safeguard against zero-weight samples. Real samples always receive weight 1.0. The weights are passed as \texttt{sample\_weight} to the classifier, so geometrically consistent synthetic samples contribute more to training.

\subsection{Baselines and Classifiers}

We compare against 8 augmentation baselines: SMOTE~\cite{chawla2002smote}, random oversampling, EDA~\cite{wei2019eda}, back-translation (via Gemini EN$\to$ES$\to$EN), no augmentation, embedding mixup, T5 paraphrasing, and BERT contextual augmentation~\cite{devlin2018bert}. Together with our two variants (binary filtering and soft weighting) these constitute the 10 methods of Table~\ref{tab:methods}, with SMOTE serving as the reference throughout. Five classifiers are evaluated: logistic regression, linear SVC, Ridge, random forest (100 trees), and MLP (100 hidden units).

\subsection{Statistical Protocol}

Experiments use up to 5 random seeds $[42, 123, 456, 789, 1011]$ controlling few-shot sampling, under two protocols. The main comparison of 10 methods (Tables~\ref{tab:methods} and~\ref{tab:nshot}) uses the 3 linear classifiers and the first 3 seeds (1,890 runs). A second protocol covers all 5 classifiers and all 5 seeds over 7 methods (3,675 runs) and supplies the MLP and random forest rows of Table~\ref{tab:classifiers}. The primary metric is macro F1 on the full test set. Significance is assessed via paired $t$-tests with Bonferroni correction; each table reports its number of paired observations $N$. Effect sizes are reported as Cohen's $d$ ($0.2$=small, $0.5$=medium, $0.8$=large). Confidence intervals are computed via bootstrap with 10,000 resamples. Win rate measures the fraction of configurations where the method exceeds SMOTE.

\section{Experimental Results}

\subsection{Main Comparison}

Table~\ref{tab:methods} presents the comparison of 10 augmentation methods against SMOTE across 1,890 configurations (21 dataset configurations $\times$ 3 classifiers $\times$ 3 seeds $\times$ 10 methods).

The soft-weighted method achieves +2.61pp over SMOTE with a large effect size ($d=0.95$) and 88.9\% win rate. Binary filtering also reaches significance (+2.39pp, $d=0.82$). Critically, all modern baselines, T5 paraphrasing, BERT contextual augmentation, and embedding mixup, \textit{fail to outperform SMOTE}. This challenges the narrative that pretrained model-based techniques automatically surpass classical methods, consistent with~\cite{cegin2025llms}.

The ablation between soft weighting and binary filtering shows a marginal difference of +0.22pp ($p=0.053$), indicating that the main contribution comes from geometric filtering itself. Binary filtering already achieves +1.75pp over no augmentation ($p<0.0001$, $d=0.63$), confirming that geometric selection adds significant value over indiscriminate sample incorporation.

\subsection{Effect of Training Set Size}
Table~\ref{tab:nshot} shows the pronounced diminishing returns pattern. At 10-shot, the effect size is very large ($d=1.48$) with absolute macro F1 improving from 67.23\% (SMOTE) to 72.12\%. At 25-shot, the win rate reaches 95.2\%, indicating near-universal improvement. At 50-shot, benefits diminish as classifiers already have sufficient data for reasonable decision boundaries. This pattern has a direct interpretation: when real data is scarce, filtered synthetic samples provide genuinely new information that densifies class regions; as real data grows, the classifier already captures the main decision boundaries and augmentation provides diminishing marginal value.

\begin{figure*}[t]
\centering
\includegraphics[scale=0.95]{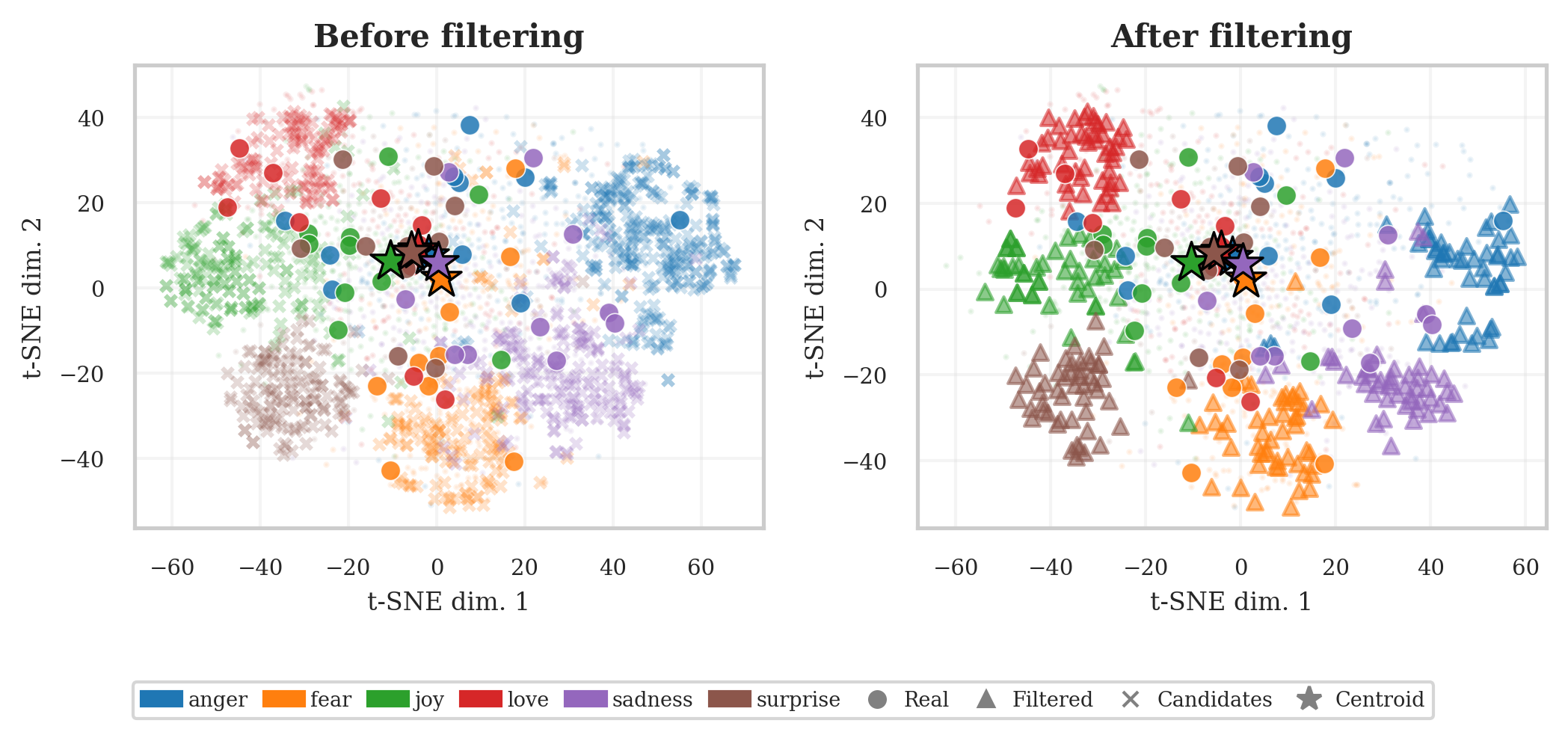}
\caption{t-SNE visualization of the embedding space for the emotion dataset (6 classes, 10-shot). Left: all LLM candidates (crosses) include samples in cross-class and peripheral regions. Right: after geometric filtering, only class-consistent samples (triangles) are retained near real examples (circles). Stars indicate class centroids.}
\label{fig:tsne}
\end{figure*}

Fig.~\ref{fig:tsne} illustrates the filtering mechanism via t-SNE projection of the emotion dataset (6 classes, 10-shot). The left panel shows all LLM candidates (crosses), many of which fall in cross-class or peripheral regions. The right panel shows only the geometrically filtered samples (triangles), which cluster tightly around the real examples (circles) of each class.

\begin{table}[t]
\centering
\caption{Comparison of 10 augmentation methods vs SMOTE baseline.}
\label{tab:methods}
\scriptsize
\setlength{\tabcolsep}{3pt}
\begin{tabular}{lrrrr}
\toprule
\textbf{Method} & \textbf{$\Delta$ (pp)} & \textbf{95\% CI} & \textbf{$d$} & \textbf{Win\%} \\
\midrule
\textbf{Soft-weighted}  & \textbf{+2.61}*** & {[+1.98, +3.25]} & \textbf{0.95} & \textbf{88.9} \\
Binary filter            & +2.39***           & {[+1.75, +3.07]} & 0.82          & 79.4          \\
No augmentation          & +0.64***           & {[+0.42, +0.89]} & 0.64          & 73.0          \\
Back-translation         & $-$0.07            & {[$-$0.42, +0.26]} & $-$0.05     & 57.1          \\
EDA                      & $-$0.15            & {[$-$0.57, +0.27]} & $-$0.08     & 52.4          \\
BERT contextual          & $-$0.26            & {[$-$0.58, +0.02]} & $-$0.21     & 46.0          \\
Emb. mixup               & $-$0.01            & {[$-$0.13, +0.11]} & $-$0.02     & 44.4          \\
Random oversample        & $-$0.16            & {[$-$0.35, +0.02]} & $-$0.21     & 39.7          \\
\textit{SMOTE (base)}    & ---               & ---                & ---           & ---           \\
T5 paraphrase            & $-$0.35            & {[$-$0.62, $-$0.11]} & $-$0.34   & 38.1          \\
\bottomrule
\multicolumn{5}{l}{\scriptsize $N\!=\!63$ paired comparisons. *** $p\!<\!0.001$ (Bonferroni).}
\end{tabular}
\end{table}

\begin{table}[t]
\centering
\caption{Performance by number of examples per class ($n$-shot).}
\label{tab:nshot}
\scriptsize
\setlength{\tabcolsep}{3pt}
\begin{tabular}{lrrr|rrr}
\toprule
 & \multicolumn{3}{c|}{\textbf{Soft-weighted}} & \multicolumn{3}{c}{\textbf{Binary filter}} \\
\textbf{$n$} & \textbf{$\Delta$} & \textbf{$d$} & \textbf{Win} & \textbf{$\Delta$} & \textbf{$d$} & \textbf{Win} \\
\midrule
10 & +4.89*** & 1.48 & 90.5 & +4.44*** & 1.15 & 87.3 \\
25 & +2.17*** & 1.57 & 95.2 & +1.96*** & 1.18 & 90.5 \\
50 & +0.76*** & 0.80 & 74.6 & +0.77*** & 0.71 & 63.5 \\
\bottomrule
\multicolumn{7}{l}{\scriptsize *** $p\!<\!0.001$ (paired $t$-test). $\Delta$ in pp.}
\end{tabular}
\end{table}

\subsection{Classifier Analysis}

Table~\ref{tab:classifiers} presents the breakdown by classifier type.

\begin{table}[t]
\centering
\caption{Performance by classifier type (soft-weighted vs SMOTE).}
\label{tab:classifiers}
\small
\begin{tabular}{lrrrr}
\toprule
\textbf{Classifier} & \textbf{$\Delta$ (pp)} & \textbf{Cohen's $d$} & \textbf{Win\%} & \textbf{$N$} \\
\midrule
Linear SVC   & +2.98*** & 1.02 & 92.1 & 63 \\
Ridge        & +2.74*** & 1.04 & 92.1 & 63 \\
MLP          & +2.16*** & 0.74 & 88.6 & 105 \\
Logistic Reg.& +2.09*** & 0.82 & 76.2 & 63 \\
Random Forest& +1.42**  & 0.35 & 61.9 & 105 \\
\bottomrule
\multicolumn{5}{l}{\footnotesize ** $p<0.01$, *** $p<0.001$ (Bonferroni-corrected).} \\
\multicolumn{5}{l}{\footnotesize $N\!=\!63$: 3-seed protocol; $N\!=\!105$: 5-seed protocol.}
\end{tabular}
\end{table}

Linear classifiers benefit most from geometric filtering. Linear SVC achieves +2.98pp ($d=1.02$) and Ridge +2.74pp ($d=1.04$), both with large effect sizes and over 92\% win rates. The MLP shows a medium-to-large effect ($d=0.74$), while random forest, although still significant, shows only a small effect ($d=0.35$), the weakest of the five. The geometric explanation is that linear classifiers establish hyperplane decision boundaries that are directly reinforced by densifying nearby regions of the embedding space. Filtered synthetic samples land in class-consistent zones close to these boundaries, providing the classifier with additional support vectors. Tree-based methods partition along individual feature axes, a strategy that does not benefit equally from spatial densification, as synthetic samples may not align with the specific axis-parallel cuts that trees use.

\subsection{Cross-Task Generalization: NER}

Table~\ref{tab:ner} presents the extension to named entity recognition. The adaptation requires no filter modification, only the definition of ``class'' changes to dominant entity type per sentence.

\begin{table}[t]
\centering
\caption{NER extension results (entity-level F1).}
\label{tab:ner}
\small
\setlength{\tabcolsep}{4pt}
\begin{tabular}{llrrrr}
\toprule
\textbf{Corpus} & \textbf{$n$} & \textbf{Baseline} & \textbf{No filter} & \textbf{Cascade} & \textbf{LOF} \\
\midrule
\multirow{3}{*}{MultiNERD}
 & 10 & .247 & .393 & .407 & \textbf{.426} \\
 & 25 & .315 & .419 & \textbf{.438} & .431 \\
 & 50 & .415 & .486 & .455 & \textbf{.495} \\
\midrule
\multirow{3}{*}{WikiANN}
 & 10 & .162 & .268 & \textbf{.303} & .302 \\
 & 25 & .253 & .272 & \textbf{.299} & .276 \\
 & 50 & .244 & \textbf{.295} & .287 & .293 \\
\midrule
\multirow{3}{*}{Few-NERD}
 & 10 & .103 & \textbf{.234} & .220 & .217 \\
 & 25 & .161 & .240 & \textbf{.251} & .235 \\
 & 50 & .200 & .248 & \textbf{.272} & .261 \\
\midrule
\multicolumn{2}{l}{\textbf{Average}} & .233 & .317 & .326 & .326 \\
\multicolumn{2}{l}{\textbf{$\Delta$ vs baseline}} & --- & +8.40pp & +9.26pp & +9.28pp \\
\bottomrule
\multicolumn{6}{l}{\footnotesize Cascade level-1: $p=0.0003$ (raw), $p_{Bonf}=0.002$ over the} \\
\multicolumn{6}{l}{\footnotesize 7 filters evaluated, $d=2.05$, 100\% win rate vs baseline.}
\end{tabular}
\end{table}

The cascade level-1 filter achieves +9.26pp over the unaugmented baseline ($p_{Bonf}=0.002$, $d=2.05$), with 100\% win rate across all 9 configurations and the best result on 5 of them. Relaxed LOF is statistically indistinguishable (+9.28pp, $d=1.91$), so the two simplest filters tie at the top, mirroring the classification results. Most of this gain is attributable to the LLM data itself: unfiltered generation already yields +8.40pp, and geometric filtering adds a further +0.86pp that does not reach significance with only 9 configurations ($p=0.27$). The pattern of diminishing returns by $n$-shot persists: +13.97pp at 10-shot, +8.67pp at 25-shot, +5.15pp at 50-shot. The combined filter produces catastrophic results with 0\% acceptance rate in 6 of 9 configurations and is the only filter failing to reach significance ($p=0.29$, $d=0.38$), reinforcing the over-filtering conclusion. This cross-task generalization confirms that distance-based filtering captures task-agnostic geometric properties of the embedding space.

\begin{figure}[t]
\centering
\includegraphics[width=\columnwidth]{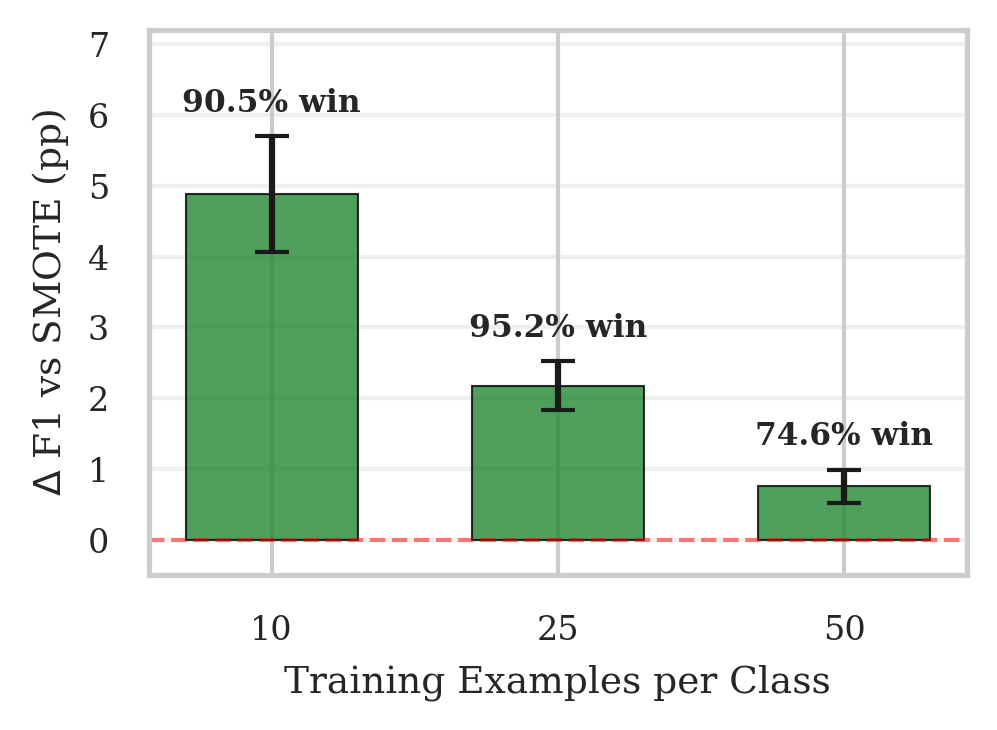}
\caption{Improvement over SMOTE ($\Delta$ in pp) by number of training examples per class. The benefit of geometric filtering decreases as more real data becomes available, confirming that the method is most valuable in extremely low-resource scenarios.}
\label{fig:nshot}
\end{figure}

Fig.~\ref{fig:nshot} shows the diminishing returns pattern for text classification, which the NER results above mirror.

\subsection{Why Simple Filters Outperform Complex Ones}

Three mechanisms explain this counterintuitive result:

\textbf{Metric correlation:} For L2-normalized vectors, Euclidean distance is a monotonic transformation of cosine similarity. Adding cosine as a separate criterion provides no new discriminative information. In 768 dimensions, cosine similarity suffers from concentration of measure~\cite{tessari2024diem}, becoming quasi-uniform.

\textbf{Diversity loss:} The combined filter's intersection of criteria eliminates diverse candidates, retaining only quasi-duplicates of existing data. Augmentation value lies in covering underrepresented regions, which restrictive intersection destroys.

\textbf{LOF instability:} With $k=20$ neighbors but only 9 real examples per class at 10-shot, LOF density estimates degenerate. Euclidean distance to the class centroid remains stable even with $N=10$ points~\cite{robinson2025manifold}. This is consistent with evidence that representation quality determines outcomes more than selection algorithm complexity~\cite{du2025disentangling}.

\section{Discussion}

\subsection{LLM Robustness}

A critical question is whether the observed gains depend on the specific LLM used for generation. Evaluation with 5 generative models, Gemini 3 Flash, GPT-5-mini, Claude 4.5 Haiku, Kimi K2.5 (open-source), and GLM-5 (open-source), shows that all achieve statistically significant improvements after Bonferroni correction for 5 comparisons. The spread between the best model (+2.31pp, Gemini) and the worst (+1.49pp, Kimi) is only 0.82pp. The Friedman test detects no significant LLM effect ($\chi^2=8.15$, $p=0.086$), and Kendall's coefficient of concordance ($W = 0.59$) indicates substantial inter-model agreement on which dataset configurations benefit most from filtering. This confirms that the geometric filtering framework is truly decoupled from the generative model.

\subsection{Per-Class Analysis}

Individual class analysis reveals that difficult classes benefit disproportionately. Classes with baseline F1 below 30\% gain +10.44pp on average, while easy classes (F1 $>$ 80\%) gain only +0.55pp (Spearman $\rho=-0.269$, which does not reach significance with the 13 classes analyzed, $p=0.37$). For example, in the emotion dataset at 10-shot, the ``anger'' class improves by +16.9pp (from 27.8\% to 44.6\%), while ``sadness'' gains just +1.0pp. This concentration of benefits where classifiers have the largest margin for improvement has practical implications: geometric filtering is especially valuable for applications where performance on minority or difficult classes is critical, such as hate speech detection or medical triage.

\subsection{Effect of Number of Classes}

The number of classes influences filtering effectiveness. Datasets with 2--20 classes show consistent positive gains, with the sweet spot at 6 classes (+5.42pp, 100\% win rate). However, with 77 classes (banking77) the result turns negative ($-0.99$pp), and with 150 classes (clinc150) it is neutral (+0.01pp). The Spearman correlation between class count and gain is $\rho = -0.260$ ($p = 0.003$). As classes multiply, each occupies a smaller embedding region, and distance-based filtering loses discriminative power when class boundaries become densely packed.

\subsection{Curriculum Learning}

An auxiliary experiment with curriculum-based inclusion of candidates ordered by geometric score reveals that the optimal inclusion threshold is approximately 50\% of candidates (+2.74pp vs SMOTE, 86.2\% win rate), surpassing inclusion of all candidates (+2.22pp) by +0.52pp. This demonstrates that roughly half of LLM-generated candidates introduce noise that degrades classifier quality, even after top-$N$ selection. The curriculum experiment provides practical guidance: generating a 3$\times$ surplus and retaining the top 50\% by geometric score maximizes augmentation value.

\subsection{Geometric Space Analysis}

Quantitative analysis of the embedding space confirms the filtering mechanism. Filtered samples produce more compact clusters (intra-class distance: 0.822 vs 0.850 unfiltered, $p<0.001$) and better-separated classes (inter-class distance: 0.616 vs 0.571, $p<0.001$). The silhouette score improves from 0.075 to 0.085 ($p=0.029$), with filtering winning in 71\% of configurations.

\subsection{Limitations}

The evaluation is English-only; multilingual transfer is unknown. Gains shrink sharply with more real data: +0.76pp at 50-shot versus +4.89pp at 10-shot, and under the 5-classifier protocol the 50-shot difference no longer reaches significance ($p=0.16$). The benefit also vanishes as classes multiply, turning negative at 77 classes ($-0.99$pp) and neutral at 150 ($+0.01$pp). The same embedding model is used for both filtering and classification, introducing potential circularity, though cross-model ablation (90.5\% agreement across 4 embedding models) and superiority over teacher-model filtering ($+2.78$ vs $+2.43$pp for the binary variants; paired geometric-vs-teacher difference $p=0.0001$) mitigate this concern. LLM response caching produces zero cross-seed variance for deterministic classifiers (80\% of configurations); restricting analysis to stochastic classifiers still yields significance (+1.79pp, $p=0.0015$, $d=0.52$).

\section{Conclusion}

We presented a geometric filtering framework for LLM-based data augmentation in few-shot text classification. The key idea is simple: generate a surplus of LLM candidates, embed them, and retain only those whose Euclidean distance to real class examples indicates geometric consistency. A soft weighting mechanism further modulates each sample's training contribution based on its filter score.

Across 6,700+ experimental configurations, 13 datasets, 5 classifiers, and 10 augmentation methods, the method achieves statistically significant improvements over SMOTE (+2.61pp, $p<0.0001$, Cohen's $d=0.95$, 88.9\% win rate). The approach generalizes to NER (+9.26pp, 100\% win rate) without filter modification, is robust across 5 LLMs from 4 providers, and concentrates its largest benefits on the
most difficult classes (+10.44pp for classes with baseline F1 $<$ 30\%).

The most practically relevant finding is that the simplest distance-based filter consistently outperforms complex multi-criteria alternatives. This counterintuitive result is explained by three mechanisms: metric correlation in L2-normalized spaces that makes cosine similarity redundant, diversity loss under restrictive filter intersection, and density estimator instability in the few-shot regime. This finding aligns with broader evidence that representation quality matters more than selection algorithm complexity~\cite{du2025disentangling}.

Future directions include multilingual evaluation with cross-lingual embedding models, validation with locally-deployed open-source LLMs (Llama, Mistral) that bypass content filters, joint Bayesian hyperparameter optimization across filter and weighting parameters, and extension to other NLP tasks such as question answering and abstractive summarization.

\section*{Acknowledgment}

This work was developed as part of the Master of Science in Data Science program at Universidad Adolfo Ib\'{a}\~{n}ez. The authors thank ANID FONDECYT 1230315, ANID-MILENIO-NCN2024\_103, ANID-MILENIO-NCN2024\_047, and Centro de Modelamiento Matemático (CMM) FB210005, BASAL funds for centers of excellence from ANID-Chile.

\bibliographystyle{IEEEtran}
\bibliography{references}

\end{document}